\documentclass{article}

 \usepackage[main, final]{neurips_2025}

\usepackage[utf8]{inputenc} 
\usepackage[T1]{fontenc}    
\usepackage{hyperref}       
\usepackage{url}            
\usepackage{booktabs}       
\usepackage{amsfonts}       
\usepackage{nicefrac}       
\usepackage{microtype}      
\usepackage{xcolor}         
\usepackage{colortbl}
\usepackage{amsmath,amssymb,amsfonts}
\usepackage{graphicx}
\usepackage{subfigure}
\usepackage{multirow}
\usepackage{listings}
\usepackage{pythonhighlight}
\usepackage{paralist}
\usepackage{arydshln}
\usepackage{pifont}
\usepackage{wrapfig}
\usepackage{caption}
\usepackage{natbib}
\usepackage{tcolorbox}
\tcbuselibrary{breakable}
\usepackage{tikz}
\definecolor{darkgreen}{RGB}{0,160,0} 

\title{Manalyzer: End-to-end Automated Meta-analysis with Multi-agent System}

\author{Wanghan Xu\textsuperscript{1,2}, Wenlong Zhang\textsuperscript{2}, Fenghua Ling\textsuperscript{2}, Ben Fei\textsuperscript{2,3},  Yusong Hu\textsuperscript{4,2}, \\ \textbf{Runmin Ma\textsuperscript{2}}, \textbf{Bo Zhang\textsuperscript{2}}, \textbf{Fangxuan Ren\textsuperscript{5}}, \textbf{Jintai Lin\textsuperscript{5}}, \textbf{Wanli Ouyang\textsuperscript{2}}, \textbf{Lei Bai\textsuperscript{2} \footnotemark[2]}
\\
\textsuperscript{1}Shanghai Jiao Tong University \quad \textsuperscript{2}Shanghai Artificial Intelligence Laboratory \\ \textsuperscript{3}The Chinese University of Hong Kong \quad \textsuperscript{4}Nankai University \quad \textsuperscript{5}Peking University
\\
\footnotemark[2] Corresponding author. \texttt{bailei@pjlab.org.cn} \\
}

\begin{document}

\maketitle

\begin{abstract}
Meta-analysis is a systematic research methodology that synthesizes data from multiple existing studies to derive comprehensive conclusions. This approach not only mitigates limitations inherent in individual studies but also facilitates novel discoveries through integrated data analysis. Traditional meta-analysis involves a complex multi-stage pipeline including literature retrieval, paper screening, and data extraction, which demands substantial human effort and time. However, while LLM-based methods can accelerate certain stages, they still face significant challenges, such as hallucinations in paper screening and data extraction. In this paper, we propose a multi-agent system, Manalyzer, which achieves end-to-end automated meta-analysis through tool calls. The hybrid review, hierarchical extraction, self-proving, and feedback checking strategies implemented in Manalyzer significantly alleviate these two  hallucinations. To comprehensively evaluate the performance of meta-analysis, we construct a new benchmark comprising 729 papers across 3 domains, encompassing text, image, and table modalities, with over 10,000 data points. Extensive experiments demonstrate that Manalyzer achieves significant performance improvements over the LLM baseline in multi meta-analysis tasks. Project page: \url{https://black-yt.github.io/meta-analysis-page/} .
\end{abstract}

\section{Introduction}


Meta-analysis~\cite{borenstein2021introduction} is a quantitative research method that systematically identifies, screens, evaluates, and synthesizes quantitative data from multiple independent studies, applying statistical methods to perform a comprehensive analysis in order to obtain a more reliable and precise pooled effect size regarding a specific research question and to reveal heterogeneity among studies and its sources, thereby enhancing statistical power and the generalizability of conclusions. This method is widely used in many scientific fields like atmospheric science~\cite{gonzalez2012meta}, agronomy~\cite{philibert2012assessment}, environmental science~\cite{mengist2020method}.

\begin{figure}[t]
\centerline
{\includegraphics[width=14cm]{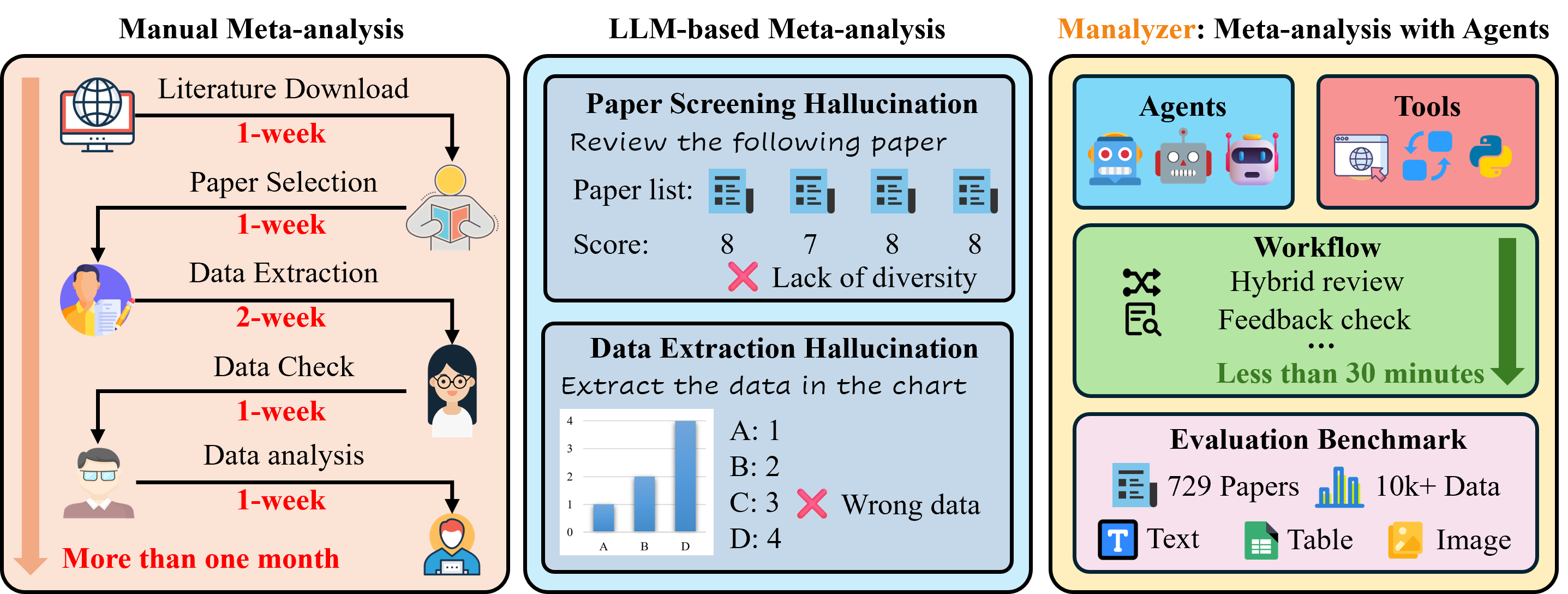}}
\caption{\textbf{Meta-analysis Comparison.} a) Manual: time-consuming. b) LLM-based: limited to specific steps, fails to achieve end-to-end automation, prone to screening and extraction hallucinations. c) Manalyzer (ours): end-to-end automation, significantly reduced hallucinations via workflow design.}
\label{fig: comparison}
\vspace{-2em}
\end{figure}

Traditional meta-analysis is a complex multi-stage, multi-task processing pipeline, which requires manual screening of hundreds of relevant papers from a massive literature library~\cite{trikalinos2008meta}, careful selection of useful data for integration~\cite{field2010meta}, and finally conclusions and reports through data analysis methods~\cite{crowther2010systematic}. This process is labor-intensive and time-consuming, often requiring the collaboration of several researchers and taking more than a month~\cite{harrison2011getting}, as shown in left subfigure of Figure~\ref{fig: comparison}.

With the advancements in large language models (LLMs)~\cite{naveed2023comprehensive}, recent works~\cite{naveed2023comprehensive, luo2024evaluating, wang2024zero} have explored leveraging LLMs to accelerate the meta-analysis pipeline, often utilizing them to assist in specific stages such as literature screening and data extraction. However, two significant hallucination issues~\cite{friel2023chainpoll} hinder the deployment of these models in real-world applications: a) LLMs tend to produce low-discriminative scores during literature screening~\cite{scherbakov2024emergence}, leading to ineffective screening processes that struggle to identify high-quality papers. b) LLMs may hallucinate during data extraction~\cite{stringhi2023hallucinating}, outputting non-existent or incorrect data, thereby compromising the reliability of the integrated data, as illustrated in the middle subfigure of Figure~\ref{fig: comparison}. These hallucination problems are challenging for a single LLM to resolve, but can be potentially mitigated by employing a multi-agent system (MAS)~\cite{dorri2018multi} leveraging direct inter-agent collaboration and supervision.

In this paper, we propose a multi-agent system named Manalyzer (\textbf{Meta-analysis analyzer}) designed to achieve end-to-end automated meta-analysis. Manalyzer integrates multiple collaborative agents and a rich toolset~\cite{gutknecht2001integrating} capable of performing sub-tasks such as literature keyword search, PDF downloading, PDF parsing, literature review, data extraction, data analysis, and report generation. 


To mitigate hallucinations in Manalyzer for the critical tasks of paper screening and data extraction, we develop several workflows~\cite{maldonado2024multi}. Specifically, we introduce a hybrid review mechanism to prevent agent score convergence, thereby reducing paper screening hallucination. This mechanism begins by conducting an individual review of each paper, generating detailed, multi-dimensional scores. Subsequently, it reviews multiple papers as a batch, enabling mutual comparison and yielding relative scores that highlight differences. By integrating these two scoring approaches, we achieve review results that are both fine-grained and diverse. Furthermore, to address the challenge of excessively long input papers, we employ a dynamic programming algorithm to extract the most valuable paragraphs, thus alleviating the context window limitation. In the data extraction stage, we design hierarchical extraction, self-proving, and feedback checking mechanisms to improve data quality.

To comprehensively evaluate the performance of LLMs and Manalyzer on meta-analysis tasks, we construct a comprehensive benchmark dataset comprising 729 academic papers across three scientific domains. This dataset encompasses data in text, table, and image modalities, and contains over 10,000 extractable data points. Experimental results demonstrate that our multi-agent system significantly outperforms the LLM baseline in the critical tasks of paper screening and data extraction.

We summarize the contributions of this paper as follows:

\begin{compactitem}
\label{sec:contributions}
\item We design a multi-agent system, Manalyzer, which implements \textbf{real-world end-to-end} meta-analysis through tool calls and significantly improves paper screening and data extraction performance via workflow designs such as hybrid review and feedback checking.

\item We introduce the first benchmark dataset in the field of scientific literature meta-analysis, comprising \textbf{10,000 data points from 729 papers} and featuring text, table, image modalities, which comprehensively evaluates the capabilities in paper screening and data extraction.

\item Experimental results show that Manalyzer significantly outperforms the LLM baselines in paper screening (\textbf{+30\% F1}) and data extraction (\textbf{+50\% hit rate}) tasks.


\end{compactitem}



\section{Related Work}

\paragraph{Meta-analysis with AI.} Meta-analysis is a method for collecting, integrating, and re-analyzing existing literature. This approach is widely applied in scientific research. For example, Root et al.~\cite{root2003fingerprints} used meta-analysis to reveal that the global average temperature has risen by approximately $0.6^{\circ}\text{C}$ over the past 100 years, quantitatively demonstrating the global warming effect. In recent years, the application of artificial intelligence in meta-analysis has gradually evolved. For instance, Luo et al.~\cite{luo2024evaluating} and Wang et al.~\cite{wang2024zero} employed LLMs to review papers and determine their inclusion in meta-analysis; Yun et al.~\cite{yun2024automatically} used prompts to guide LLMs in extracting tabular data from medical clinical reports; Torres et al.~\cite{torres2024promptheus} applied BERTopic~\cite{grootendorst2022bertopic} for topic modeling in meta-analysis and utilized the T5 model~\cite{ni2021sentence} for paper summarization; Reason et al.~\cite{reason2024artificial} leveraged LLMs to summarize multiple papers and generate comprehensive reports; Ahad et al.~\cite{ahad2024empowering} automated the literature search and screening process by integrating retrieval-augmented generation (RAG)~\cite{lewis2020retrieval}, followed by summarizing papers and generating reports. However, most existing studies focus only on specific aspects of meta-analysis rather than the complete workflow. Furthermore, the application of multi-agent systems based on tool calling in meta-analysis remains underexplored.

\paragraph{Multi-agent Systems for Scientific Research.} Multi-agent systems (MAS)~\cite{dorri2018multi} involve collaborating LLMs or VLMs for complex tasks. Recently, more MAS aim to accelerate scientific research. For instance, Ghafarollahi et al.~\cite{ghafarollahi2024sciagents} introduced dynamic collaboration among LLM-powered agents to perform knowledge retrieval, protein structure analysis, physics-based simulation, and result analysis, providing a versatile solution for protein design and analysis problems. Zheng et al.~\cite{zheng2023chatgpt} employed agents to conduct experimental design, code editing, and robotic operations in chemical research, significantly improving the efficiency of material synthesis experiments. Beyond these domain-specific MAS, some general-purpose MAS can also support scientific research. For example, Deep Research~\cite{jesudason2025openai} and Manus~\cite{hughes2025ai} demonstrate strong general task processing capabilities through web search and tool invocation. However, the number of papers they can search and process is insufficient for meta-analysis, which typically requires handling hundreds of papers.

In the field of meta-analysis, there is not only a lack of specialized MAS but also standardized, large-scale evaluation benchmarks. Benchmarks such as Humanity's Last Exam~\cite{phan2025humanity} and GAIA~\cite{mialon2023gaia} assess MAS capabilities in scientific domains through a single problem-solving paradigm, making them unsuitable for meta-analysis tasks. Therefore, this work fills this research gap by proposing a meta-analysis MAS with a comprehensive evaluation benchmark.

\begin{table}[ht]
\vspace{-1.5em}
\caption{\textbf{Comparison of Meta-analysis Systems across Key Dimensions:} (1) End-to-end workflow coverage, (2) Multi-agent architecture, (3) Specialized benchmark, (4) Tool calling capability, (5) Feedback mechanisms, (6) Large-scale literature processing, and (7) Real world application. The comparison highlights Manalyzer's comprehensive capabilities in automated meta-analysis.}
\vspace{+0.4em}
\centering
\resizebox{14.2cm}{!}{
\renewcommand{\arraystretch}{1.3}
\begin{tabular}{l|ccccccc}
\toprule
\textbf{System} & \textbf{End-to-end} & \textbf{Multi-agent} & \textbf{Benchmark} & \textbf{Tool Calling} & \textbf{Feedback} & \textbf{Large-scale} & \textbf{Real-world}\\
\hline
Luo et al.~\cite{luo2024evaluating} & \color{red}{\texttimes} & \color{red}{\texttimes} & \color{red}{\texttimes} & \color{red}{\texttimes} & \color{red}{\texttimes} & \color{red}{\texttimes} & \color{red}{\texttimes}\\
Wang et al.~\cite{wang2024zero} & \color{red}{\texttimes} & \color{red}{\texttimes} & \color{red}{\texttimes} & \color{red}{\texttimes} & \color{red}{\texttimes} & \color{red}{\texttimes} & \color{red}{\texttimes}\\
Yun et al.~\cite{yun2024automatically} & \color{red}{\texttimes} & \color{red}{\texttimes} & \color{red}{\texttimes} & \color{darkgreen}{\checkmark} & \color{red}{\texttimes} & \color{red}{\texttimes} & \color{red}{\texttimes}\\
Torres et al.~\cite{torres2024promptheus} & \color{red}{\texttimes} & \color{red}{\texttimes} & \color{red}{\texttimes} & \color{darkgreen}{\checkmark} & \color{red}{\texttimes} & \color{red}{\texttimes} & \color{red}{\texttimes}\\
Ahad et al.~\cite{ahad2024empowering} & \color{darkgreen}{\checkmark} & \color{red}{\texttimes} & \color{red}{\texttimes} & \color{darkgreen}{\checkmark} & \color{red}{\texttimes} & \color{darkgreen}{\checkmark} & \color{red}{\texttimes}\\
\hdashline[3pt/4pt]
Deep Research~\cite{jesudason2025openai} & \color{darkgreen}{\checkmark} & \color{darkgreen}{\checkmark} & \color{red}{\texttimes} & \color{darkgreen}{\checkmark} & \color{darkgreen}{\checkmark} & \color{red}{\texttimes} & \color{darkgreen}{\checkmark}\\
Manus~\cite{hughes2025ai} & \color{darkgreen}{\checkmark} & \color{darkgreen}{\checkmark} & \color{red}{\texttimes} & \color{darkgreen}{\checkmark} & \color{darkgreen}{\checkmark}  & \color{red}{\texttimes}  & \color{darkgreen}{\checkmark}\\
\hdashline[3pt/4pt]
\rowcolor{blue!15}\textbf{Manalyzer (Ours)} & \color{darkgreen}{\checkmark} & \color{darkgreen}{\checkmark} & \color{darkgreen}{\checkmark} & \color{darkgreen}{\checkmark} & \color{darkgreen}{\checkmark} & \color{darkgreen}{\checkmark} & \color{darkgreen}{\checkmark}\\
\bottomrule
\end{tabular}
}
\label{tab:meta-analysis-comparison}
\end{table}

\begin{figure}[t]
\centerline
{\includegraphics[width=15cm]{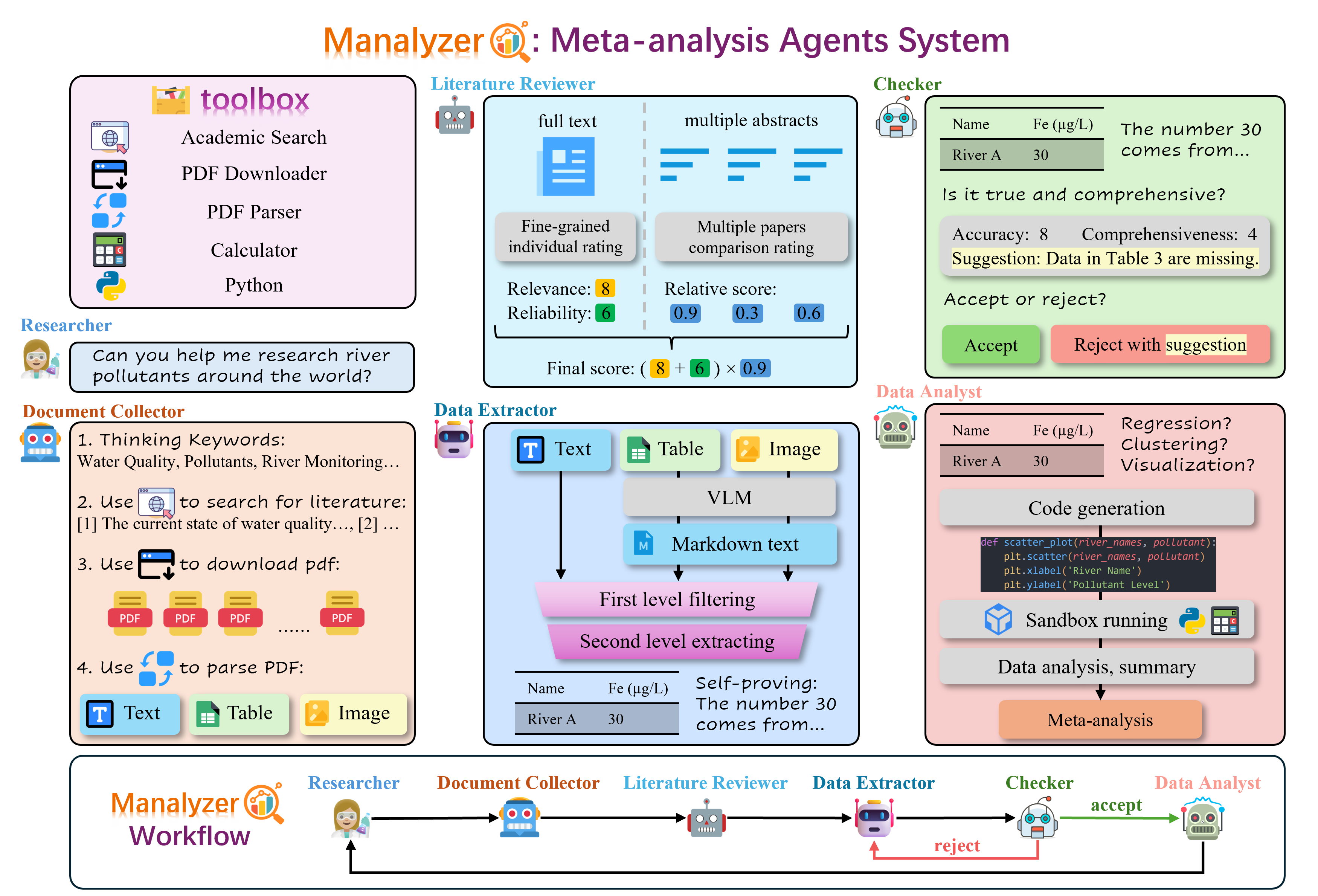}}
\caption{\textbf{Overview of Manalyzer.} Manalyzer uses multi-agent collaboration with tools to automate the full meta-analysis workflow: search, download, parsing, data extraction, data analysis.}
\label{fig: Framework}
\vspace{-2em}
\end{figure}

\section{Manalyzer: Meta-analysis with Multi-agent System}

\paragraph{Overview.} Manalyzer is a multi-agent system incorporating tool calling and feedback mechanisms, enabling end-to-end automated meta-analysis in real scientific research scenarios. We divide the meta-analysis process into three stages. The first stage involves receiving user input, searching for and downloading papers, followed by filtering out relevant and valuable ones. The second stage focuses on extracting data from these selected papers and integrating it into tables. The third stage is to analyze the integrated data and output the final meta-analysis report.

\subsection{Stage 1: Paper Searching, Downloading, Screening}

\paragraph{Document Collector.} In this stage, the user first inputs the research direction. The keyword generator (implemented by LLM) generates a combination of multiple keywords based on the research direction input by the user. The paper downloader searches for a large number of relevant papers by calling the search API of the academic platform, obtains paper information such as paper title, paper doi number~\cite{liu2021digital}, etc., and attempts to download the PDF. 

Upon acquiring the PDF of the research paper, the PDF parser initiates an Optical Character Recognition (OCR)-based tool~\cite{islam2017survey}, such as MinerU~\cite{wang2024mineru}, to scrutinize the PDF content. The parser subsequently outputs three lists: a text list \( L_{\text{tx}} \), a figure list \( L_{\text{fg}} \), and a table list \( L_{\text{tb}} \). Each element in the text list \( L_{\text{tx}} \) corresponds to a paragraph in the paper, while the figure list \( L_{\text{fg}} \) and table list \( L_{\text{tb}} \) store the figures and tables from the paper, respectively, along with their corresponding captions.

\begin{wrapfigure}[14]{r}{0.5\textwidth}
    \vspace{-16pt}
    \includegraphics[scale=0.53, trim={0 0 0 0}, clip]{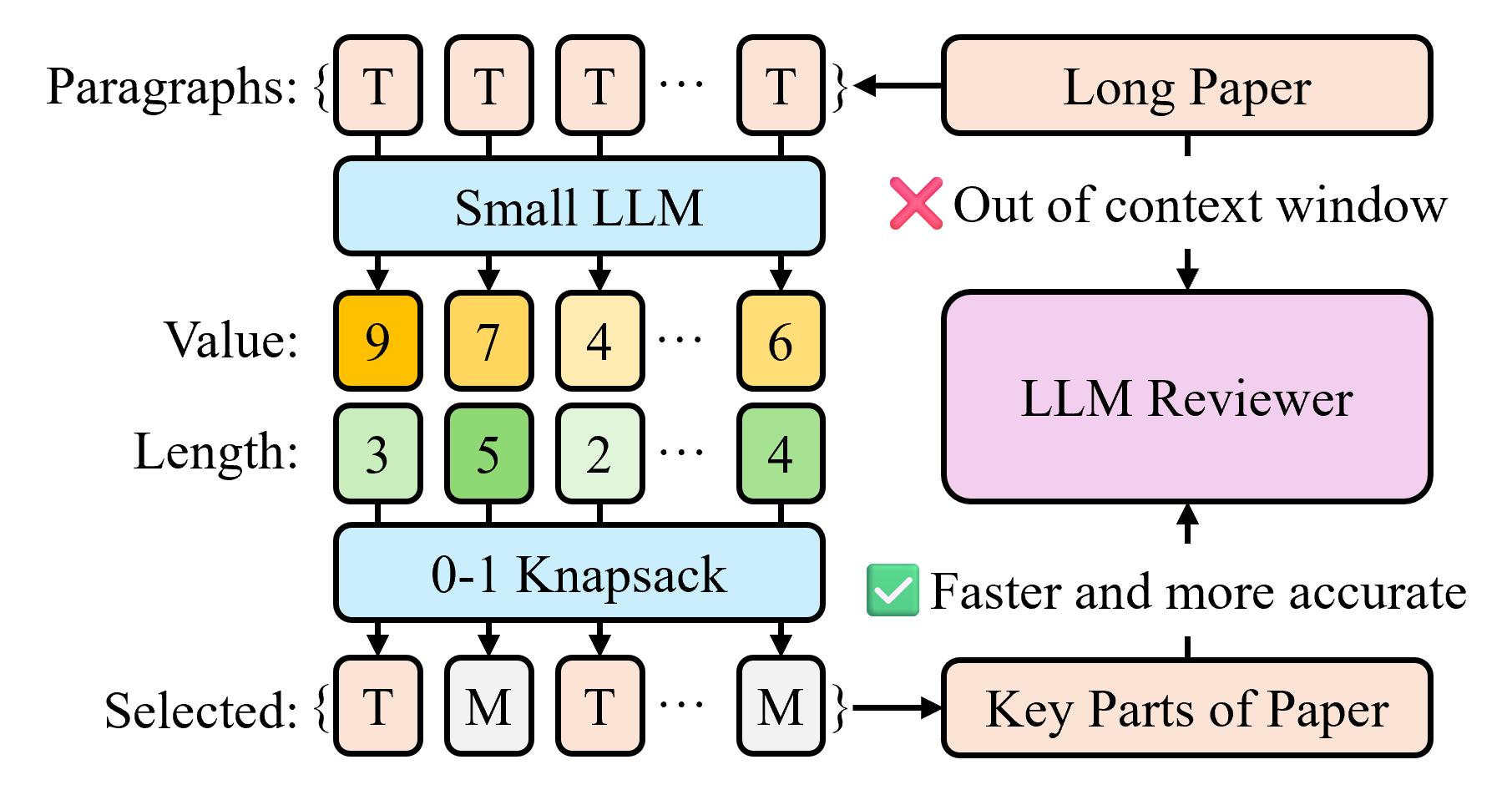}
    \vspace{-5pt}
    \caption{\textbf{Long Paper Review.} Use the knapsack algorithm to address the issue of long papers exceeding the context window limit of LLMs.}
    \label{fig:01}
\end{wrapfigure}

\paragraph{Literature Reviewer with Hybrid Review Mechanism.} Subsequently, the paper reviewers score each paper on two dimensions: data relevance ($s_1$) and data reliability ($s_2$). This step, termed independent review, allows for fine-grained, multi-dimensional scoring by processing the full paper. To handle context limits~\cite{ding2024longrope}, we dynamically select the most informative paragraphs for the model. First, a small LLM rates each paragraph's importance. Then, a knapsack-like dynamic programming algorithm~\cite{martello1987algorithms} chooses a paragraph set maximizing total importance within the length constraint (Figure~\ref{fig:01}). The reviewer model then scores these paragraphs on relevance and reliability.

Independent review can yield similar scores, hindering effective paper screening. To address this hallucination, we propose a \textbf{hybrid review}: after obtaining independent review scores for each paper, papers are batched ($n=20$) for cross-comparison, yielding relative score ($s_r$, ranging from 0 to 1). Comparison highlights the strengths and weaknesses of papers more clearly, resulting in a wider distribution of $s_r$. Finally, we calculate the final score for each paper as $s_r \times (s_1 + s_2)$, combining fine-grained assessment with relative standing. By setting a final score threshold, we filter a subset of papers from the original collection for subsequent process.

\subsection{Stage 2: Data Extraction}

 

\paragraph{Data Extractor with Hierarchical Extraction and Self-proving Mechanism.} For both table images ($L_{\text{tb}}$) and potentially tabular figures ($L_{\text{fg}}$), we employ a VLM~\cite{zhang2024vision} to generate Markdown-formatted tables~\cite{gruber2012markdown}, leveraging captions as context. To ensure clarity in subsequent data extraction, particularly with abbreviations, the VLM provides a two-level description: a summary of the main content and a detailed footnote of each row and column. Figures not suitable for table conversion (e.g., diagrams) are processed by the VLM to extract their key information as bullet points.

Following this initial processing, the resulting text, figures, and tables are batched and input as text into the extractor implemented using an LLM. The extractor performs data extraction through a \textbf{hierarchical approach}. Initially, it generates a binary mask (0 or 1) for each input part, indicating whether it contains valid data relevant to the meta-analysis theme. This efficiently filters out irrelevant information. Subsequently, only the sections identified as containing valid data are fed back into the extractor, which is then prompted to output the relevant data in a Markdown table format ($T_{\text{et}}$).

To mitigate hallucinations during this second extraction phase, we implemented a \textbf{self-proving} strategy. This requires the extractor to provide evidence for each numerical value in the output table by citing its origin in the original text. This significantly reduces the likelihood of the extractor generating non-existent data, and the provided proofs can be used for subsequent validation.

\paragraph{Checker with Feedback Mechanism.} While the self-proving strategy effectively mitigates model hallucinations, we further introduce a dedicated checker to evaluate the reasonableness and correctness of the extractor's outputs. This checker takes as input the raw data from the paper (text, figures, tables) and the integrated table ( $T_{\text{et}}$ ) generated by the extractor. It outputs scores for accuracy (verifying the correctness of values) and consistency (ensuring data semantics align with the thematic requirements), along with modification suggestions. In cases of low scores, these suggestions are fed back to the extractor as revision prompts. This feedback loop iterates until the extractor produces an integrated table ( $T_{\text{et}}$ ) with accurate values that satisfy the thematic requirements.

\subsection{Stage 3: Data Analysis and Report Output}

\paragraph{Data Analyst with Code Generation.} Manalyzer is a universal framework applicable to any disciplinary domain. Due to significant differences in the format and content of data across disciplines, fixed data analysis methods are insufficient to effectively handle the open-ended data space. Therefore, we designed a code generation-based~\cite{gu2023llm} data analysis approach. Specifically, the data analysis module first generates diverse data analysis code, such as clustering~\cite{rokach2005clustering}, classification~\cite{novakovic2017evaluation}, and regression~\cite{nunez2011regression}, based on the collection of tables extracted from all papers \( \{T_{\text{et},i}\} \), where \( i \) represents the paper index. Subsequently, the code is executed in a sandbox environment, reading the real data from \( \{T_{\text{et},i}\} \) and saving the results in the form of visualization images or tables. 

As the concluding step of the meta-analysis, the reporter (implemented by a VLM) takes the user-defined topic, the collection of extracted data tables \( \{T_{\text{et},i}\} \), and the data analysis results as input to generate a comprehensive report. This report encompasses details regarding data sources, data distribution, and analytical insights derived from the data.

\section{Evaluation Benchmark}

\paragraph{Overview.} To comprehensively and objectively evaluate the performance of Manalyzer and LLM baselines in meta-analysis, we introduce the first benchmark dataset derived from real-world and large-scale scientific papers. Given that paper screening and data extraction are the most time-consuming and labor-intensive stages in manual meta-analysis, we will use these as our evaluation tasks.

\subsection{Task 1: Paper Screening}

\paragraph{Data Construction.} High-quality papers are key for valid meta-analysis. Thus, we established a paper screening task to assess LLM or MAS ability in selecting quality papers. We chose "PM 2.5 pollutant content in China from 2003 to 2014" as the research focus for paper screening and downloaded 182 related academic papers from the internet as the initial paper collection. Human experts in the field of atmospheric science reviewed each paper in the initial collection and comprehensively judged its suitability for subsequent data capture and analysis based on the following two aspects:

\begin{itemize}
\item \textbf{Data Relevance}: Whether the data within the paper meets the specific requirements of the research focus (e.g., Korea PM2.5 or China 2015-2020 PM2.5 are irrelevant for this topic).
\item \textbf{Data Reliability}: Whether the data in the paper originates from a well-designed experiment, meets statistical significance requirements, and includes a complete textual description. 
\end{itemize}

\paragraph{Task Definition.} By considering both data relevance and data reliability, domain experts ultimately determined whether a paper could be used in the subsequent data extraction and data analysis processes. Consequently, 69 papers from the initial collection were labeled as usable. LLMs and MAS are required to emulate human experts by analyzing each paper in the initial collection to determine its usability, and their judgments will then be compared against human labels.



\subsection{Task 2: Data Extraction}

\paragraph{Data Construction.} Data extraction is the core of literature meta-analysis. This process involves extracting useful data from dozens to hundreds of academic papers according to the research directions and integrating them into a unified table. We deconstruct three manually completed meta-analysis studies~\cite{ren2023evaluation, pittelkow2015productivity, kumar2019global} from the research fields of atmosphere, agriculture, and environment, and compile a dataset of 729 academic papers with over 10,000 data points, as shown in Table~\ref{tab:paper source}.

\begin{table}[ht]
\vspace{-1.em}
\caption{\textbf{Data Distribution in Extraction Task.} The benchmark includes 729 papers with 10,000+ data points across three fields, which assesses models extract research-relevant data from multimodal content (tables, images, text) and consolidate it into structured tables.}
\vspace{+0.4em}
\resizebox{14cm}{!}{
\centering
\renewcommand{\arraystretch}{1.3}
\begin{tabular}{l|cccccc}
\toprule
\textbf{Field} & \textbf{Research Direction} & \textbf{\#Papers} & \textbf{\#Tables} & \textbf{\#Images} & \textbf{\#Extract Data}  \\
\hline
Atmosphere~\cite{ren2023evaluation}   & The PM2.5 levels in different regions of China. & 111 & 331 & 754 & 1,030  \\
Agriculture~\cite{pittelkow2015productivity}  & Production of crops in various regions of the world. & 507 & 2,452 & 1377 & 9,082  \\
Environment~\cite{kumar2019global}  & Heavy metal content in rivers around the world. & 111 & 553 & 461 & 1,330  \\
\hdashline[3pt/4pt]
Total        &                                                 & \textbf{729} & \textbf{3,336} & \textbf{2,592} & \textbf{11,442}  \\
\bottomrule
\end{tabular}
}
\label{tab:paper source}
\vspace{-1.em}
\end{table}

\paragraph{Task Definition.} To better evaluate the data extraction capabilities of different models, we established three distinct evaluation levels based on the actual application scenarios of the tasks:

\begin{itemize}
\vspace{-0.5em}
\item \textbf{Level 1}: Extracting data from paper text. Textual data provides ample context and generally lacks numerical reading errors, making it relatively easy to extract.
\item \textbf{Level 2}: Extracting data from paper tables and images. The extraction process from tables and images can lead to hallucinations, such as generating wrong values or non-existent ones.
\item \textbf{Level 3}: Obtaining data through calculations. Some data cannot be directly obtained from the paper but require operations like unit conversion, summation, and averaging. This necessitates the model's understanding of data semantics and computational ability.
\vspace{-0.5em}

\end{itemize}

\label{hit rate}
LLMs and MAS are required to extract data relevant to the research direction from each paper (including text, tables and images). Subsequently, their extractions are compared against human-extracted data to calculate the hit rate. Specifically, for each paper, $N_1$ denotes the set of data identified by the model, and $N_2$ represents the set of data identified by human experts . The hit rate is calculated as ( $|N_1 \cap N_2| / |N_2|$ ), where $|\cdot|$ represents the cardinality of the set.

\section{Experiment}

In this section, we evaluate the performance of Manalyzer and LLM baselines on two core tasks of meta-analysis, paper screening and data extraction, using the benchmark we constructed.

\subsection{Experimental Setup}
\label{Experimental Setup}

For Task 1, we select 4 open-source and 6 closed-source LLMs as baselines. We use prompts to instruct the models to score each paper on two dimensions: Relevance ($s_1$, ranging from 1 to 10) and Reliability ($s_2$, ranging from 1 to 10). Relevance assesses the paper's alignment with the research direction, while Reliability focuses on the trustworthiness of the data within the paper. Finally, we apply a threshold of $(s_1 + s_2) / 2 > 6$ as the screening criterion to obtain the paper screening results. For Manalyzer, we employ our proposed hybrid review mechanism for paper screening. We use Accuracy (Acc.), Precision (Pre.), Recall (Rec.), and F1-score as evaluation metrics.

For Task 2, we select 4 open-source and 7 closed-source VLMs as baselines. We use prompts to instruct the models to extract data relevant to the research direction from text, tables, and images, and to output this data in Markdown table format. For Manalyzer, we utilize hierarchical extraction, self-proving, and feedback checking mechanisms for data extraction. We use hit rate defined in Section~\ref{hit rate} as the evaluation metric. All models are tested with a temperature setting of 0.

\subsection{Task 1: Paper Screening}
\label{Task 1: Paper Screening}
\begin{figure}[!ht]

\centering
\resizebox{5.8cm}{!}{
\begin{minipage}[h]{0.43\textwidth}
\centering
\captionof{table}{\textbf{Classification skills of different models in screening papers.}}
\label{tab:f1}
\begin{tabular}{l|p{0.5cm}p{0.5cm}p{0.5cm}p{0.5cm}}
\toprule
\textbf{Model} & \textbf{Acc.}$\uparrow$ & \textbf{Pre.}$\uparrow$ & \textbf{Rec.}$\uparrow$ & \textbf{F1}$\uparrow$ \\
\midrule
Llama-3-8B & 43.4 & 41.7 & 100 & 58.8 \\
Llama-3-70B & 43.3 & 40.0 & 98.5 & 56.9 \\
Qwen-2.5-72B & 47.4 & 47.5 & 100 & 64.3 \\
DeepSeek-V3 & 39.0 & 38.3 & 100 & 55.4 \\
\hdashline[3pt/4pt]
GPT-4o & 60.9 & 49.2 & 88.4 & 63.2 \\
Gemini-1.5 & 45.6 & 45.1 & 100 & 62.1 \\
Gemini-2.5 & 39.2 & 38.1 & 98.5 & 54.9 \\
Grok-3 & 38.4 & 37.9 & 98.5 & 54.8 \\
Claude-3.5 & 37.3 & 37.5 & 98.5 & 54.4 \\
Claude-3.7 & 38.7 & 37.2 & 98.4 & 54.0 \\
\rowcolor{blue!15}\textbf{Manalyzer} & \textbf{80.8} & \textbf{70.7} & 84.1 & \textbf{76.8} \\
\bottomrule
\end{tabular}
\end{minipage}
}
\hfill
\resizebox{8.0cm}{!}{
\begin{minipage}[h]{0.55\textwidth}
\centering
\includegraphics[width=1.0\linewidth]{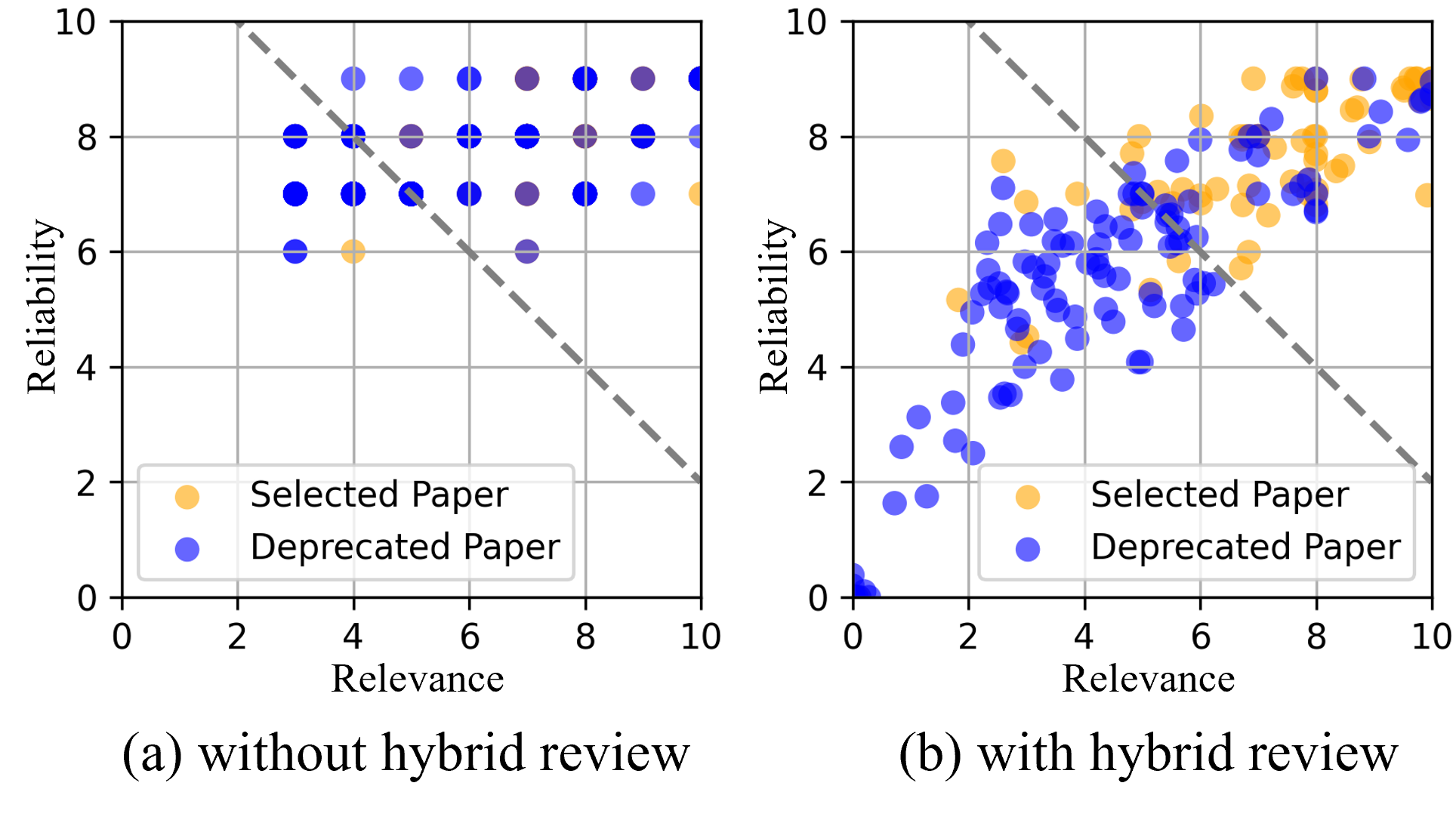}
\vspace{-0.6cm}
\captionof{figure}{\textbf{Distribution of Paper Scores under Different Review Strategies.} Hybrid review strategy improves score diversity and selects more suitable papers.}
\label{fig:score diversity}
\end{minipage}
}
\end{figure}

Table~\ref{tab:f1} presents the performance of different models in paper screening. The results indicate that simple reasoning paradigms fail to significantly distinguish the quality of papers. By analyzing the scoring results of the models (as shown in Figure~\ref{fig:score diversity} (a)), the score distribution of GPT-4 for the 182 papers in the original dataset is highly concentrated, leading to a lack of differentiation in the scores, termed paper screening hallucination. The fundamental reason for this phenomenon lies in individually reviewing papers, which cannot unify evaluation criteria or create gaps between samples through comparative analysis. Additionally, the models exhibit a clear tendency to over-praise, meaning that even if a paper has certain issues, the model may still assign a high score. To address this issue, the hybrid review strategy proposed in this paper is necessary.

Compared to these models, Manalyzer shows significant improvements in both accuracy and recall. Figure~\ref{fig:score diversity} (b) shows the score distribution of Manalyzer for the 182 papers, which is more dispersed and exhibits higher differentiation. This indicates that the second stage of the hybrid review strategy better reflects the relative quality of each paper in the original dataset by introducing comparisons among papers, thereby facilitating the selection of higher-quality papers for subsequent data extraction.

\subsection{Task 2: Data Extraction}

\begin{table}[t]
\vspace{-1em}
\caption{\textbf{Hit Rate of Data Extraction.} The benchmark encompasses 3 domains, with 3 increasing levels of difficulty. Level 1 involves extracting numbers from text, Level 2 focuses on extracting numbers from tables and images, and Level 3 entails extracting numbers that require calculation.}
\vspace{+0.4em}
\resizebox{14cm}{!}{
\centering
\renewcommand{\arraystretch}{1.3}
\begin{tabular}{l|lll|lll|lll}
\toprule
& \multicolumn{3}{c}{Atmosphere} & \multicolumn{3}{c}{Agriculture} & \multicolumn{3}{c}{Environment}\\
\textbf{Model} & \textbf{level 1} & \textbf{level 2} & \textbf{level 3} & \textbf{level 1} & \textbf{level 2} & \textbf{level 3} & \textbf{level 1} & \textbf{level 2} & \textbf{level 3} \\
\hline
Llama-3.2-11B-V~\cite{grattafiori2024llama} &62.77     &42.32     &\textbf{4.35}     &24.84     &18.50     &1.21     &36.51     &42.63     &0.19     \\
Qwen-2.5-32~\cite{yang2024qwen2} &21.28     &29.79     &0.00     &33.54     &29.88     &7.91     &46.03     &46.01     &2.27     \\
Qwen-2.5-72B~\cite{yang2024qwen2} &53.09     &52.84     &0.00     &34.62     &30.48     &10.07     &55.56     &59.13     &3.26     \\
DeepSeek-VL2~\cite{wu2024deepseek} &30.85     &20.21     &0.00     &18.00     &20.62     &0.35     &17.46     &29.60     &0.00     \\
\hdashline[3pt/4pt]
GPT-4o~\cite{islam2024gpt} &44.68     &45.05     &0.00     &24.22     &22.12     &7.25     &42.86     &43.84     &0.76     \\
GPT-4-V~\cite{zhang2023gpt} &59.57     &45.05     &0.16     &34.78     &27.81     &5.89     &47.62     &37.35     &0.57     \\
Gemini-1.5~\cite{team2024gemini} &56.38     &59.73     &0.16     &37.27     &39.02     &17.56     &42.86     &43.03     &1.14     \\
Gemini-2.5~\cite{team2023gemini} &63.83     &53.92     &0.93     &02.48     &6.87     &2.93     &4.76     &5.82     &0.00     \\
Grok-3~\cite{de2025grok} &60.64     &45.05     &0.47     &19.88     &2.11     &2.40     &31.75     &7.31     &1.33     \\
Claude-3.5~\cite{kurokawa2024diagnostic} &62.77     &55.63     &0.00     &40.37     &47.75     &4.60     &69.84     &50.74     &4.73     \\
Claude-3.7~\cite{lim2025evaluating} &63.83     &58.70     &1.09     &40.37     &43.32     &18.31     &53.97     &41.27     &2.46     \\
\rowcolor{blue!15}\textbf{Manalyzer (Ours)} &\textbf{77.66}     &\textbf{70.65}     &3.42     &\textbf{60.84}     &\textbf{64.37}     &\textbf{30.46}     &\textbf{72.88}     &\textbf{63.37}     &\textbf{5.23}     \\
\bottomrule
\end{tabular}
}
\label{tab:main-results}
\end{table}

Table~\ref{tab:main-results} presents the data extraction hit rates of different VLMs (using various reasoning methods) and Manalyzer under different task difficulty levels. The results show that most models struggle to accurately extract the target data. By analyzing specific cases of model extraction, we observe that: on the one hand, most models extract incomplete data, often with significant omissions, which is particularly evident when dealing with large tables; on the other hand, some models exhibit consistent errors in data extraction, such as extracting the content of nitrogen dioxide instead of sulfur dioxide as required. This phenomenon is also referred to as data extraction hallucination.

To alleviate this issue, our hierarchical extraction strategy decouples the data extraction process into two parts: the first step uses a 01 mask to filter out irrelevant data, and the second step refines the data extraction. This decoupling reduces the difficulty of each step and improves the coverage of data extraction. Through the self-proving strategy, Manalyzer significantly reduces the generation of incorrect or non-existent data because this data cannot be proven. Finally, the feedback checker examines the data extraction results and provides modification suggestions, which can further reduce data omissions or errors. More analysis is presented in the ablation study~\ref{ablation}.

\subsection{Case Studies}

\begin{figure}[t]
\centerline
{\includegraphics[width=14cm]{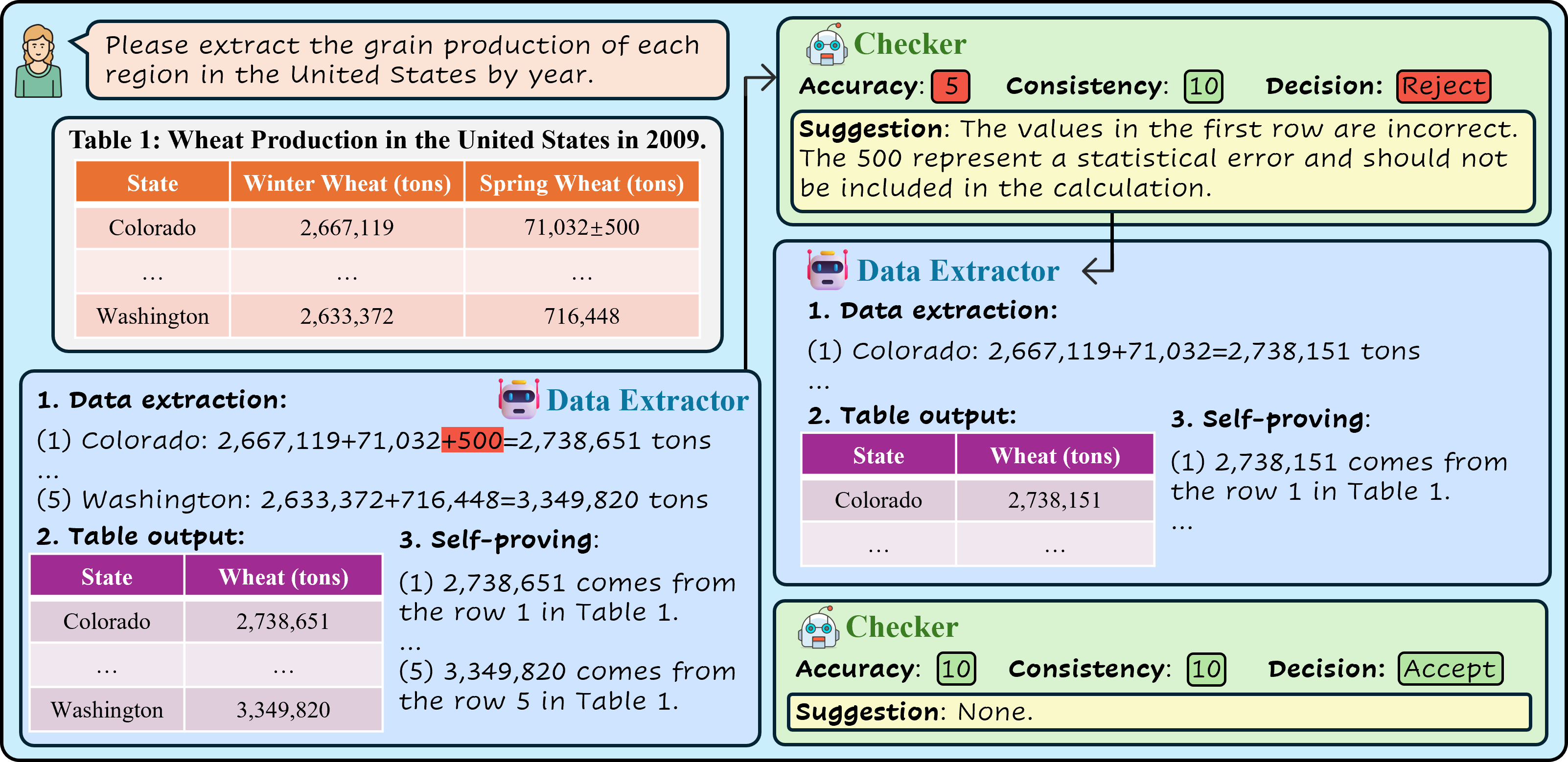}}
\caption{\textbf{Data Extraction Case.} This is a Level 3 example as data extraction requires computation. The example demonstrates the roles of Self-proving and the Checker.}
\label{fig: case}
\end{figure}

Figure~\ref{fig: case} illustrates the roles of self-proving and the checker during data extraction. This is a level 3 extraction case, meaning the target data is not directly found in the table and requires computation during extraction. In this example, simple addition is needed to obtain the total value of wheat production. Similar calculation processes include averaging, etc. Level 3 data extraction requires the model to have semantic understanding of the data and numerical computation abilities, making it relatively difficult during data extraction. Consequently, Level 3 hit rates are generally low in Table~\ref{tab:main-results}.

During data extraction, the checker, as an independent agent, supervises the data extractor and provides corresponding suggestions. If the checker rejects a result, the process reverts to the data extractor, and the suggestions serve as additional input to guide more accurate extraction. In actual experiments, the probability of checker feedback being triggered is approximately 12\%, with the majority of rejections resolved within one cycle. We set the maximum feedback loop to 3 times.



\subsection{Ablation Studies}
\label{ablation}
Ablation study investigates hybrid review's impact on paper screening accuracy and hierarchical extraction, self-proving, feedback checker's effect on Manalyzer data extraction hit rate.

\begin{figure}[!ht]
\centering
\begin{minipage}[h]{0.45\textwidth}
\centering
\includegraphics[width=0.8\linewidth]{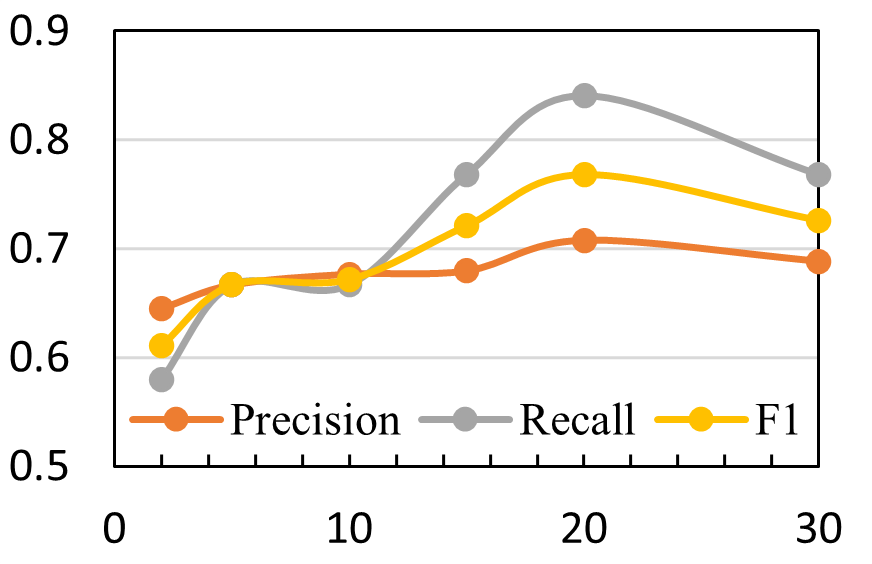}
\vspace{-0.2cm}
\captionof{figure}{\textbf{Impact of Batch Size on Paper Screening Metrics.} Metrics increases first and then decreases as the batch size increases.}
\label{fig:batch_review}
\vspace{-0.2cm}
\end{minipage}
\hfill
\resizebox{7.6cm}{!}{
\begin{minipage}[h]{0.56\textwidth}
\centering
\captionof{table}{\textbf{Impact of Different Components of Manalyzer on Hit Rate.} The hierarchical extraction improves accuracy by pre-filtering irrelevant data, self-proving reduces hallucinations during data extraction, and feedback minimizes omissions via external verification.}
\label{tab:ablation}
\begin{tabular}{l|ccc}
\toprule
\textbf{Method} & \textbf{Level 1} & \textbf{Level 2} & \textbf{Level 3} \\
\midrule
Baseline & 44.6 & 45.0 & 0 \\
\ +Hierarchical extract & 65.9 & 55.3 & 0.5 \\
\ \ +Self-proving & 71.3 & 67.9 & 1.1 \\
\rowcolor{blue!15}\ \ \ +Feedback checker & \textbf{77.7} & \textbf{70.6} & \textbf{3.4} \\
\bottomrule
\end{tabular}
\end{minipage}
}

\end{figure}


\paragraph{Hybrid Review.} Figures~\ref{fig:score diversity} (a) and (b) show score distributions for individual and hybrid review. Individual review scores are concentrated, while hybrid review scores are dispersed, aiding paper distinction. Figure~\ref{fig:batch_review} demonstrates the impact of different batch sizes on score dispersion and paper screening accuracy. When the batch size is 0, the system reverts to single-paper review. As the batch size increases, the paper screening accuracy first rises and then declines. This is because a too-small batch size increases randomness within the batch, while a too-large batch size makes it more difficult for the model to score simultaneously. Thus, we select a batch size of 20 as optimal.

\paragraph{Hierarchical Extraction.} Hierarchical extraction is a two-step data extraction process. First, it determines each section's relevance to the meta-analysis theme, performing an initial screening of tables, figures, and paragraphs. Subsequently, it extracts specific numerical data from the filtered content. This decoupling transforms the original "selective extraction" task into two simpler sub-tasks. The first sub-task focuses solely on initial screening for relevance, while the second sub-task concentrates on precise data extraction from the relevant sections. This significantly simplifies each stage and enhances data extraction coverage, as evidenced by the improved hit rates in Table~\ref{tab:ablation}.

\paragraph{Self-proving.} Self-proving refers to the process where the data extractor provides the specific source of the data alongside the extracted data. This method effectively prevents the model from generating false or non-existent data during extraction. Table~\ref{tab:ablation} shows the proportion of extracted data that does not match the original text after applying self-proving. The experimental results demonstrate that self-proving significantly reduces the occurrence of hallucinations in large models.
 
\paragraph{Feedback Mechanism.} The feedback mechanism improves the accuracy and consistency of data extraction by introducing an independent checker agent to identify issues during the process. Given that hierarchical extraction and self-proving already significantly improve the quality of numerical extraction, the checker's feedback mechanism is not triggered in every instance. The results in Table~\ref{tab:ablation} demonstrate that the checker enhances the quality of the data extraction process. 

\section{Conclusion}

Meta-analysis, a crucial methodology for synthesizing findings across studies, traditionally demands significant human effort in its multi-stage pipeline. While LLMs offer potential for acceleration, challenges such as hallucinations in paper screening and data extraction persist. To address these limitations, this paper introduces Manalyzer, a multi-agent system designed for end-to-end automated meta-analysis leveraging tool calls. Manalyzer incorporates key strategies, including hybrid review for robust paper screening and hierarchical extraction with self-proving and feedback checking for accurate data extraction. These mechanisms significantly alleviate the issues of hallucinations in both critical stages. To comprehensively evaluate meta-analysis performance, we present a new benchmark dataset comprising 729 papers across three diverse domains, featuring text, image, and table modalities and containing over 10,000 data points. Extensive experiments on this benchmark demonstrate the significant performance gains achieved by Manalyzer over LLM baselines in both paper screening and data extraction tasks, highlighting the effectiveness of the proposed multi-agent approach and the introduced hallucination mitigation strategies for automated meta-analysis.

The limitation of this work lies in the absence of benchmarks for evaluating aspects such as paper downloads, analysis, and report outputs, as these elements are difficult to quantify.



\clearpage

\bibliographystyle{plain}
\bibliography{references.bib}

@book{borenstein2021introduction,
  title={Introduction to meta-analysis},
  author={Borenstein, Michael and Hedges, Larry V and Higgins, Julian PT and Rothstein, Hannah R},
  year={2021},
  publisher={John wiley \& sons}
}

@article{gonzalez2012meta,
  title={Meta-analysis on atmospheric carbon capture in Spain through the use of conservation agriculture},
  author={Gonz{\'a}lez-S{\'a}nchez, EJ and Ord{\'o}{\~n}ez-Fern{\'a}ndez, R and Carbonell-Bojollo, R and Veroz-Gonz{\'a}lez, O and Gil-Ribes, JA},
  journal={Soil and Tillage Research},
  volume={122},
  pages={52--60},
  year={2012},
  publisher={Elsevier}
}

@article{philibert2012assessment,
  title={Assessment of the quality of meta-analysis in agronomy},
  author={Philibert, Aurore and Loyce, Chantal and Makowski, David},
  journal={Agriculture, Ecosystems \& Environment},
  volume={148},
  pages={72--82},
  year={2012},
  publisher={Elsevier}
}

@article{mengist2020method,
  title={Method for conducting systematic literature review and meta-analysis for environmental science research},
  author={Mengist, Wondimagegn and Soromessa, Teshome and Legese, Gudina},
  journal={MethodsX},
  volume={7},
  pages={100777},
  year={2020},
  publisher={Elsevier}
}

@article{trikalinos2008meta,
  title={Meta-analysis methods},
  author={Trikalinos, Thomas A and Salanti, Georgia and Zintzaras, Elias and Ioannidis, John PA},
  journal={Advances in genetics},
  volume={60},
  pages={311--334},
  year={2008},
  publisher={Elsevier}
}

@article{field2010meta,
  title={How to do a meta-analysis},
  author={Field, Andy P and Gillett, Raphael},
  journal={British Journal of Mathematical and Statistical Psychology},
  volume={63},
  number={3},
  pages={665--694},
  year={2010},
  publisher={Wiley Online Library}
}

@article{crowther2010systematic,
  title={Systematic review and meta-analysis methodology},
  author={Crowther, Mark and Lim, Wendy and Crowther, Mark A},
  journal={Blood, The Journal of the American Society of Hematology},
  volume={116},
  number={17},
  pages={3140--3146},
  year={2010},
  publisher={American Society of Hematology Washington, DC}
}

@article{harrison2011getting,
  title={Getting started with meta-analysis},
  author={Harrison, Freya},
  journal={Methods in Ecology and Evolution},
  volume={2},
  number={1},
  pages={1--10},
  year={2011},
  publisher={Wiley Online Library}
}

@article{naveed2023comprehensive,
  title={A comprehensive overview of large language models},
  author={Naveed, Humza and Khan, Asad Ullah and Qiu, Shi and Saqib, Muhammad and Anwar, Saeed and Usman, Muhammad and Akhtar, Naveed and Barnes, Nick and Mian, Ajmal},
  journal={arXiv preprint arXiv:2307.06435},
  year={2023}
}

@article{luo2024evaluating,
  title={Evaluating the Efficacy of Large Language Models for Systematic Review and Meta-Analysis Screening},
  author={Luo, Ronald and Sastimoglu, Ziya and Faisal, Abu Ilius and Deen, M Jamal},
  journal={medRxiv},
  pages={2024--06},
  year={2024},
  publisher={Cold Spring Harbor Laboratory Press}
}

@inproceedings{wang2024zero,
  title={Zero-shot generative large language models for systematic review screening automation},
  author={Wang, Shuai and Scells, Harrisen and Zhuang, Shengyao and Potthast, Martin and Koopman, Bevan and Zuccon, Guido},
  booktitle={European Conference on Information Retrieval},
  pages={403--420},
  year={2024},
  organization={Springer}
}

@article{friel2023chainpoll,
  title={Chainpoll: A high efficacy method for llm hallucination detection},
  author={Friel, Robert and Sanyal, Atindriyo},
  journal={arXiv preprint arXiv:2310.18344},
  year={2023}
}

@article{scherbakov2024emergence,
  title={The emergence of Large Language Models (LLM) as a tool in literature reviews: an LLM automated systematic review},
  author={Scherbakov, Dmitry and Hubig, Nina and Jansari, Vinita and Bakumenko, Alexander and Lenert, Leslie A},
  journal={arXiv preprint arXiv:2409.04600},
  year={2024}
}

@article{stringhi2023hallucinating,
  title={Hallucinating (or poorly fed) LLMs? The problem of data accuracy},
  author={Stringhi, Elisabetta},
  journal={i-lex},
  volume={16},
  number={2},
  pages={54--63},
  year={2023}
}

@article{dorri2018multi,
  title={Multi-agent systems: A survey},
  author={Dorri, Ali and Kanhere, Salil S and Jurdak, Raja},
  journal={Ieee Access},
  volume={6},
  pages={28573--28593},
  year={2018},
  publisher={IEEE}
}

@inproceedings{gutknecht2001integrating,
  title={Integrating tools and infrastructures for generic multi-agent systems},
  author={Gutknecht, Olivier and Ferber, Jacques and Michel, Fabien},
  booktitle={Proceedings of the fifth international conference on Autonomous agents},
  pages={441--448},
  year={2001}
}

@article{maldonado2024multi,
  title={Multi-agent Systems: A survey about its components, framework and workflow},
  author={Maldonado, Diego and Cruz, Edison and Torres, Jackeline Abad and Cruz, Patricio J and Gamboa, Silvana},
  journal={IEEE Access},
  year={2024},
  publisher={IEEE}
}

@article{root2003fingerprints,
  title={Fingerprints of global warming on wild animals and plants},
  author={Root, Terry L and Price, Jeff T and Hall, Kimberly R and Schneider, Stephen H and Rosenzweig, Cynthia and Pounds, J Alan},
  journal={Nature},
  volume={421},
  number={6918},
  pages={57--60},
  year={2003},
  publisher={Nature Publishing Group UK London}
}

@article{yun2024automatically,
  title={Automatically extracting numerical results from randomized controlled trials with large language models},
  author={Yun, Hye Sun and Pogrebitskiy, David and Marshall, Iain J and Wallace, Byron C},
  journal={arXiv preprint arXiv:2405.01686},
  year={2024}
}

@article{torres2024promptheus,
  title={PROMPTHEUS: A Human-Centered Pipeline to Streamline SLRs with LLMs},
  author={Torres, Jo{\~a}o Pedro Fernandes and Mulligan, Catherine and Jorge, Joaquim and Moreira, Catarina},
  journal={arXiv preprint arXiv:2410.15978},
  year={2024}
}

@article{grootendorst2022bertopic,
  title={BERTopic: Neural topic modeling with a class-based TF-IDF procedure},
  author={Grootendorst, Maarten},
  journal={arXiv preprint arXiv:2203.05794},
  year={2022}
}

@article{ni2021sentence,
  title={Sentence-t5: Scalable sentence encoders from pre-trained text-to-text models},
  author={Ni, Jianmo and Abrego, Gustavo Hernandez and Constant, Noah and Ma, Ji and Hall, Keith B and Cer, Daniel and Yang, Yinfei},
  journal={arXiv preprint arXiv:2108.08877},
  year={2021}
}

@article{reason2024artificial,
  title={Artificial intelligence to automate network meta-analyses: Four case studies to evaluate the potential application of large language models},
  author={Reason, Tim and Benbow, Emma and Langham, Julia and Gimblett, Andy and Klijn, Sven L and Malcolm, Bill},
  journal={PharmacoEconomics-Open},
  volume={8},
  number={2},
  pages={205--220},
  year={2024},
  publisher={Springer}
}

@article{lewis2020retrieval,
  title={Retrieval-augmented generation for knowledge-intensive nlp tasks},
  author={Lewis, Patrick and Perez, Ethan and Piktus, Aleksandra and Petroni, Fabio and Karpukhin, Vladimir and Goyal, Naman and K{\"u}ttler, Heinrich and Lewis, Mike and Yih, Wen-tau and Rockt{\"a}schel, Tim and others},
  journal={Advances in neural information processing systems},
  volume={33},
  pages={9459--9474},
  year={2020}
}

@inproceedings{ahad2024empowering,
  title={Empowering Meta-Analysis: Leveraging Large Language Models for Scientific Synthesis},
  author={Ahad, Jawad Ibn and Sultan, Rafeed Mohammad and Kaikobad, Abraham and Rahman, Fuad and Amin, Mohammad Ruhul and Mohammed, Nabeel and Rahman, Shafin},
  booktitle={2024 IEEE International Conference on Big Data (BigData)},
  pages={1615--1624},
  year={2024},
  organization={IEEE}
}

@article{ghafarollahi2024sciagents,
  title={Sciagents: Automating scientific discovery through multi-agent intelligent graph reasoning},
  author={Ghafarollahi, Alireza and Buehler, Markus J},
  journal={arXiv preprint arXiv:2409.05556},
  year={2024}
}

@article{zheng2023chatgpt,
  title={ChatGPT chemistry assistant for text mining and the prediction of MOF synthesis},
  author={Zheng, Zhiling and Zhang, Oufan and Borgs, Christian and Chayes, Jennifer T and Yaghi, Omar M},
  journal={Journal of the American Chemical Society},
  volume={145},
  number={32},
  pages={18048--18062},
  year={2023},
  publisher={ACS Publications}
}

@article{jesudason2025openai,
  title={OpenAI's ‘Deep Research’for the Generation of Comprehensive Referenced Medical Text: Uses and Cautions},
  author={Jesudason, Daniel and Gao, Christina and Seth, Ishith and Chan, Weng Onn and Bacchi, Stephen},
  journal={ANZ Journal of Surgery},
  year={2025},
  publisher={John Wiley \& Sons Australia, Ltd Melbourne}
}

@article{hughes2025ai,
  title={AI Agents and Agentic Systems: A Multi-Expert Analysis},
  author={Hughes, Laurie and Dwivedi, Yogesh K and Malik, Tegwen and Shawosh, Mazen and Albashrawi, Mousa Ahmed and Jeon, Il and Dutot, Vincent and Appanderanda, Mandanna and Crick, Tom and De’, Rahul and others},
  journal={Journal of Computer Information Systems},
  pages={1--29},
  year={2025},
  publisher={Taylor \& Francis}
}

@article{phan2025humanity,
  title={Humanity's Last Exam},
  author={Phan, Long and Gatti, Alice and Han, Ziwen and Li, Nathaniel and Hu, Josephina and Zhang, Hugh and Zhang, Chen Bo Calvin and Shaaban, Mohamed and Ling, John and Shi, Sean and others},
  journal={arXiv preprint arXiv:2501.14249},
  year={2025}
}

@inproceedings{mialon2023gaia,
  title={Gaia: a benchmark for general ai assistants},
  author={Mialon, Gr{\'e}goire and Fourrier, Cl{\'e}mentine and Wolf, Thomas and LeCun, Yann and Scialom, Thomas},
  booktitle={The Twelfth International Conference on Learning Representations},
  year={2023}
}

@article{liu2021digital,
  title={Digital object identifier (DOI) and DOI services: An overview},
  author={Liu, Jia},
  journal={Libri},
  volume={71},
  number={4},
  pages={349--360},
  year={2021},
  publisher={De Gruyter}
}

@article{islam2017survey,
  title={A survey on optical character recognition system},
  author={Islam, Noman and Islam, Zeeshan and Noor, Nazia},
  journal={arXiv preprint arXiv:1710.05703},
  year={2017}
}

@article{wang2024mineru,
  title={Mineru: An open-source solution for precise document content extraction},
  author={Wang, Bin and Xu, Chao and Zhao, Xiaomeng and Ouyang, Linke and Wu, Fan and Zhao, Zhiyuan and Xu, Rui and Liu, Kaiwen and Qu, Yuan and Shang, Fukai and others},
  journal={arXiv preprint arXiv:2409.18839},
  year={2024}
}

@article{martello1987algorithms,
  title={Algorithms for knapsack problems},
  author={Martello, Silvano and Toth, Paolo},
  journal={North-Holland Mathematics Studies},
  volume={132},
  pages={213--257},
  year={1987},
  publisher={Elsevier}
}

@article{ding2024longrope,
  title={Longrope: Extending llm context window beyond 2 million tokens},
  author={Ding, Yiran and Zhang, Li Lyna and Zhang, Chengruidong and Xu, Yuanyuan and Shang, Ning and Xu, Jiahang and Yang, Fan and Yang, Mao},
  journal={arXiv preprint arXiv:2402.13753},
  year={2024}
}

@article{zhang2024vision,
  title={Vision-language models for vision tasks: A survey},
  author={Zhang, Jingyi and Huang, Jiaxing and Jin, Sheng and Lu, Shijian},
  journal={IEEE Transactions on Pattern Analysis and Machine Intelligence},
  year={2024},
  publisher={IEEE}
}

@inproceedings{gu2023llm,
  title={Llm-based code generation method for golang compiler testing},
  author={Gu, Qiuhan},
  booktitle={Proceedings of the 31st ACM Joint European Software Engineering Conference and Symposium on the Foundations of Software Engineering},
  pages={2201--2203},
  year={2023}
}

@article{ren2023evaluation,
  title={Evaluation of CMIP6 model simulations of PM 2.5 and its components over China},
  author={Ren, Fangxuan and Lin, Jintai and Xu, Chenghao and Adeniran, Jamiu A and Wang, Jingxu and Martin, Randall V and van Donkelaar, Aaron and Hammer, Melanie and Horowitz, Larry and Turnock, Steven T and others},
  journal={EGUsphere},
  volume={2023},
  pages={1--26},
  year={2023},
  publisher={Copernicus Publications G{\"o}ttingen, Germany}
}

@article{pittelkow2015productivity,
  title={Productivity limits and potentials of the principles of conservation agriculture},
  author={Pittelkow, Cameron M and Liang, Xinqiang and Linquist, Bruce A and Van Groenigen, Kees Jan and Lee, Juhwan and Lundy, Mark E and Van Gestel, Natasja and Six, Johan and Venterea, Rodney T and Van Kessel, Chris},
  journal={Nature},
  volume={517},
  number={7534},
  pages={365--368},
  year={2015},
  publisher={Nature Publishing Group UK London}
}

@article{kumar2019global,
  title={Global evaluation of heavy metal content in surface water bodies: A meta-analysis using heavy metal pollution indices and multivariate statistical analyses},
  author={Kumar, Vinod and Parihar, Ripu Daman and Sharma, Anket and Bakshi, Palak and Sidhu, Gagan Preet Singh and Bali, Aditi Shreeya and Karaouzas, Ioannis and Bhardwaj, Renu and Thukral, Ashwani Kumar and Gyasi-Agyei, Yeboah and others},
  journal={Chemosphere},
  volume={236},
  pages={124364},
  year={2019},
  publisher={Elsevier}
}

@article{grattafiori2024llama,
  title={The llama 3 herd of models},
  author={Grattafiori, Aaron and Dubey, Abhimanyu and Jauhri, Abhinav and Pandey, Abhinav and Kadian, Abhishek and Al-Dahle, Ahmad and Letman, Aiesha and Mathur, Akhil and Schelten, Alan and Vaughan, Alex and others},
  journal={arXiv preprint arXiv:2407.21783},
  year={2024}
}

@article{yang2024qwen2,
  title={Qwen2. 5 technical report},
  author={Yang, An and Yang, Baosong and Zhang, Beichen and Hui, Binyuan and Zheng, Bo and Yu, Bowen and Li, Chengyuan and Liu, Dayiheng and Huang, Fei and Wei, Haoran and others},
  journal={arXiv preprint arXiv:2412.15115},
  year={2024}
}

@article{wu2024deepseek,
  title={Deepseek-vl2: Mixture-of-experts vision-language models for advanced multimodal understanding},
  author={Wu, Zhiyu and Chen, Xiaokang and Pan, Zizheng and Liu, Xingchao and Liu, Wen and Dai, Damai and Gao, Huazuo and Ma, Yiyang and Wu, Chengyue and Wang, Bingxuan and others},
  journal={arXiv preprint arXiv:2412.10302},
  year={2024}
}

@article{islam2024gpt,
  title={Gpt-4o: The cutting-edge advancement in multimodal llm},
  author={Islam, Raisa and Moushi, Owana Marzia},
  journal={Authorea Preprints},
  year={2024},
  publisher={Authorea}
}

@article{zhang2023gpt,
  title={Gpt-4v (ision) as a generalist evaluator for vision-language tasks},
  author={Zhang, Xinlu and Lu, Yujie and Wang, Weizhi and Yan, An and Yan, Jun and Qin, Lianke and Wang, Heng and Yan, Xifeng and Wang, William Yang and Petzold, Linda Ruth},
  journal={arXiv preprint arXiv:2311.01361},
  year={2023}
}

@article{team2024gemini,
  title={Gemini 1.5: Unlocking multimodal understanding across millions of tokens of context},
  author={Team, Gemini and Georgiev, Petko and Lei, Ving Ian and Burnell, Ryan and Bai, Libin and Gulati, Anmol and Tanzer, Garrett and Vincent, Damien and Pan, Zhufeng and Wang, Shibo and others},
  journal={arXiv preprint arXiv:2403.05530},
  year={2024}
}

@article{team2023gemini,
  title={Gemini: a family of highly capable multimodal models},
  author={Team, Gemini and Anil, Rohan and Borgeaud, Sebastian and Alayrac, Jean-Baptiste and Yu, Jiahui and Soricut, Radu and Schalkwyk, Johan and Dai, Andrew M and Hauth, Anja and Millican, Katie and others},
  journal={arXiv preprint arXiv:2312.11805},
  year={2023}
}

@article{kurokawa2024diagnostic,
  title={Diagnostic performances of Claude 3 Opus and Claude 3.5 Sonnet from patient history and key images in Radiology’s “Diagnosis Please” cases},
  author={Kurokawa, Ryo and Ohizumi, Yuji and Kanzawa, Jun and Kurokawa, Mariko and Sonoda, Yuki and Nakamura, Yuta and Kiguchi, Takao and Gonoi, Wataru and Abe, Osamu},
  journal={Japanese Journal of Radiology},
  pages={1--4},
  year={2024},
  publisher={Springer}
}

@article{lim2025evaluating,
  title={Evaluating the Efficacy of Large Language Models in Generating Medical Documentation: A Comparative Study of ChatGPT-4, ChatGPT-4o, and Claude},
  author={Lim, Bryan and Seth, Ishith and Maxwell, Molly and Cuomo, Roberto and Ross, Richard J and Rozen, Warren M},
  journal={Aesthetic Plastic Surgery},
  pages={1--12},
  year={2025},
  publisher={Springer}
}

@article{de2025grok,
  title={Grok, Gemini, ChatGPT and DeepSeek: Comparison and Applications in Conversational Artificial Intelligence},
  author={de Carvalho Souza, Murillo Edson and Weigang, Li},
  journal={INTELIGENCIA ARTIFICIAL},
  volume={2},
  number={1},
  year={2025}
}

@article{gruber2012markdown,
  title={Markdown: Syntax},
  author={Gruber, John},
  journal={URL http://daringfireball. net/projects/markdown/syntax. Retrieved on June},
  volume={24},
  pages={640},
  year={2012}
}

@article{rokach2005clustering,
  title={Clustering methods},
  author={Rokach, Lior and Maimon, Oded},
  journal={Data mining and knowledge discovery handbook},
  pages={321--352},
  year={2005},
  publisher={Springer}
}

@article{novakovic2017evaluation,
  title={Evaluation of classification models in machine learning},
  author={Novakovi{\'c}, Jasmina Dj and Veljovi{\'c}, Alempije and Ili{\'c}, Sini{\v{s}}a S and Papi{\'c}, {\v{Z}}eljko and Tomovi{\'c}, Milica},
  journal={Theory and Applications of Mathematics \& Computer Science},
  volume={7},
  number={1},
  pages={39},
  year={2017},
  publisher={" Aurel Vlaicu" University of Arad Department of Mathematics and Computer~…}
}

@article{nunez2011regression,
  title={Regression modeling strategies},
  author={Nunez, Eduardo and Steyerberg, Ewout W and Nunez, Julio},
  journal={Revista Espa{\~n}ola de Cardiolog{\'\i}a (English Edition)},
  volume={64},
  number={6},
  pages={501--507},
  year={2011},
  publisher={Elsevier}
}

@article{hendricks2020crossref,
  title={Crossref: The sustainable source of community-owned scholarly metadata},
  author={Hendricks, Ginny and Tkaczyk, Dominika and Lin, Jennifer and Feeney, Patricia},
  journal={Quantitative Science Studies},
  volume={1},
  number={1},
  pages={414--427},
  year={2020},
  publisher={MIT Press One Rogers Street, Cambridge, MA 02142-1209, USA journals-info~…}
}

@article{ginsparg2011arxiv,
  title={ArXiv at 20},
  author={Ginsparg, Paul},
  journal={Nature},
  volume={476},
  number={7359},
  pages={145--147},
  year={2011},
  publisher={Nature Publishing Group UK London}
}







\clearpage
\appendix

\section{Document Collector}

The primary responsibility of the document collector is to search for and download PDFs. During the search process, multiple sets of keywords are generated based on user input, which facilitates searching on academic platforms to find relevant literature. The prompt for providing keywords is as follows:

\begin{tcolorbox}[breakable,title=Prompt for Keyword Search]

\textbf{\textcolor{blue}{System Prompt}}\\ 
You are an expert academic researcher in <INPUT1> with extensive experience in literature search and systematic reviews. When given a research topic or area of interest, generate a comprehensive set of search keywords organized by conceptual categories. 

\textbf{\textcolor{blue}{Instructions}}\\ 
\textbf{For each input topic}:

1. Analyze the core concepts and related subfields

2. Identify technical terms, synonyms, and variant phrasings

3. Include broader and narrower terms to ensure search flexibility

4. Group keywords thematically into logical clusters

\textbf{Output format requirements}:

- Provide only a list of lists (no explanations or headers)

- Each sublist should contain closely related terms

- Include 15-30 total keywords for most topics

- Order terms from most to least central within each group

- Use standardized academic terminology

\textbf{\textcolor{blue}{Example}}\\ 
Example for "neural networks in medical imaging":

[["Deep Learning", "Convolutional Neural Networks", "CNN", "AI Diagnostics"], 

["Medical Imaging", "Radiology", "MRI", "CT Scan", "Ultrasound"],

["Image Segmentation", "Feature Extraction", "Classification", "Computer-Aided Diagnosis"]]

\end{tcolorbox}
\begin{figure}[ht]
    \vspace{0.01cm}
    \caption{\textbf{Prompt for Keyword Search.}}
    \label{fig:Prompt for Keyword Search}
\end{figure}

Two APIs are used for literature searches: the first is CrossRef~\cite{hendricks2020crossref}, and the second is arXiv~\cite{ginsparg2011arxiv}. For the former, the search results include DOI numbers, which we use to download PDFs. For the latter, the search results contain direct download links from arXiv, which we use to obtain the PDFs.

For PDF parsing, we use the MinerU~\cite{wang2024mineru} tool, which automates the conversion of PDFs into plain text and images. This facilitates subsequent data reading and processing.

\section{Literature Reviewer}

The responsibility of the reviewing agent is to read papers and identify the strongest sections. Experiments in Section~\ref{Task 1: Paper Screening} indicate that reviewing each paper individually leads to similar scores for most papers, making differentiation difficult. To address this issue, we propose a hybrid review method that combines individual reviews of each paper with relative scores derived from comparisons between papers. This approach significantly enhances the ability to distinguish between the papers, as shown in Figure~\ref{fig:score diversity}.

During the individual review phase, we use the following prompt to evaluate each paper from multiple dimensions.

\begin{tcolorbox}[breakable,title=Prompt for Individual Review]

\textbf{\textcolor{blue}{System Prompt}}\\ 
You are a professional reviewer, your professional field is <INPUT1>, please review the following paper and rate it.

\textbf{\textcolor{blue}{Instructions}}\\ 
You'll need to assess the response on the following dimensions: Topic Relevance and Feasibility.
Evaluate the paper on different dimensions, pointing out its strengths or weaknesses in each dimension and assigning a score of 1 to 10 for each.

In general, the higher the quality of the paper and the more closely it follows the user requirements, the higher the score will be. Papers that do not meet the user requirements will receive lower scores.

\textbf{\textcolor{blue}{Scoring Rules}}\\ 
\textbf{Topic Relevance}

Scores 1-2 when the paper is not relevant to the user's needs.

Scores 3-4 when the paper belongs to the same field as the topic that the user is interested in, but there is no direct connection.

Scores 5-6 when the paper is related to the topic that the user is concerned about, but does not meet the specific requirements (such as time, place, method).

Scores 7-8 when the paper is closely related to the topic of interest to the user and meets most requirements (such as time, location, and method).

Scores 9-10 when the paper is strongly related to the topic that the user is interested in and meets all requirements (such as time, location, and method).

\textbf{Feasibility}

Scores 1-2 when the paper is not supported by any experiments or data.

Scores 3-4 when the paper has little experimental or data support.

Scores 5-6 when there are some experiments and data in the paper, but they are not complete and sufficient.

Scores 7-8 when the paper provides sufficient experiments and data, and is reproducible.

Scores 9-10 when the experiments and data in the paper are very complete, including experimental details and data descriptions, which can serve as the basis for subsequent work.

\end{tcolorbox}
\begin{figure}[ht]
    \vspace{0.01cm}
    \caption{\textbf{Prompt for Individual Review.}}
    \label{fig:Prompt for Individual Review}
\end{figure}

During the full-text review, some longer papers may exceed the context length of the LLM. Therefore, we use a smaller LLM to score each paragraph of lengthy papers, determining the value of each paragraph. Subsequently, using the 0/1 knapsack algorithm, we can identify the most valuable paragraphs that meet the length constraints as input, as shown in Figure~\ref{fig:01}. When scoring each paragraph, we use the following prompt.

\begin{tcolorbox}[breakable,title=Prompt for Paragraph Scoring]

\textbf{\textcolor{blue}{System Prompt}}\\ 
Please give the following paragraph a score to indicate the value of academic analysis. 

\textbf{\textcolor{blue}{Instructions}}\\ 
If the content of a paragraph is all general descriptive text, it is considered low value. 

Otherwise, it is given a high score. The score must be an integer from 0 to 10. 

You only need to give a single number without any other text.

\textbf{\textcolor{blue}{Example}}\\ 
8

\end{tcolorbox}
\begin{figure}[ht]
    \vspace{0.01cm}
    \caption{\textbf{Prompt for Paragraph Scoring.}}
    \label{fig:Prompt for Paragraph Scoring}
\end{figure}

After completing the individual reviews, we also conduct comparative reviews to provide differentiated scoring. By comparing papers against each other, we can obtain more diverse scores. During the comparative review process, we use the following prompt:

\begin{tcolorbox}[breakable,title=Prompt for Comparative Review]

\textbf{\textcolor{blue}{System Prompt}}\\ 
You are an expert in the field of <INPUT1> and are good at literature analysis.
Please judge whether the following papers are relevant to the user's topic of interest.

\textbf{\textcolor{blue}{Instructions}}\\ 
1. For each paper, please use a real number between 0 and 1 to indicate whether it is relevant to the topic of interest, where 1 means very relevant and 0 means completely irrelevant. Please do not give other responses.

2. When judging whether it is relevant, please follow every requirement put forward by the user in the topic of interest (including location, time, etc.).

3. Please express the final answer in the form of a list, and do not use other words to respond.

\textbf{\textcolor{blue}{Example}}\\ 
$[0.8, 0.9, 0.6, 0.1, ...]$

\end{tcolorbox}
\begin{figure}[ht]
    \vspace{0.01cm}
    \caption{\textbf{Prompt for Comparative Review.}}
    \label{fig:Prompt for Comparative Review}
\end{figure}

\section{Data Extractor}

The data extractor needs to first convert the tables and images in the paper into Markdown format text. After that, the data can be extracted and integrated. During the conversion to text, we use the following prompt:

\begin{tcolorbox}[breakable,title=Prompt for Table and Image Conversion]

\textbf{\textcolor{blue}{System Prompt}}\\ 
You are an expert in the field of <INPUT1>, and you are good at converting the table in the paper from the image format to markdown text. Please convert the following jpg table into markdown text.

\textbf{\textcolor{blue}{Instructions}}\\ 
1. Ensure that the converted data matches the original data in both values and units, clearly indicating the units.

2. Preserve the original format of the table.

3. If there are multiple tables in the image, convert each one separately.

4. For each table, give its title to reflect the content of the table.

5. For each table, provide a footnote describing the full name of each row and column.

\textbf{\textcolor{blue}{Example}}\\ 
```markdown

| Date       | Precipitation (mm)  | Type          |

|------------|---------------------|---------------|

| 2023-01-01 | 5.0                 | Rain          |

| 2023-01-02 | 12.3                | Rain          |

| 2023-01-03 | 0.0                 | None          |

| 2023-01-04 | 8.5                 | Rain          |

| 2023-01-05 | 15.0                | Snow/Rain Mix |

```\\

[The Start of Title]

Precipitation Records in the New York Area. 

[The End of Title]\\

[The Start of Footnote]

Date: The date when the precipitation was recorded, in the format of YYYY-MM-DD (year-month-day).

Precipitation (mm): The amount of precipitation on this date, in millimeters. The amount of precipitation reflects the intensity of the precipitation.

Type: The type of precipitation, describing the nature of the precipitation on this date, such as rain, snow, or mixed precipitation.

[The End of Footnote]

\end{tcolorbox}
\begin{figure}[ht]
    \vspace{0.01cm}
    \caption{\textbf{Prompt for Table and Image Conversion.}}
    \label{fig:Prompt for Table and Image Conversion}
\end{figure}

Footnotes help improve consistency during the data integration phase, reducing ambiguities caused by differences in variable abbreviations and preventing the incorrect merging of data.

After the data conversion, we extract user-relevant data from the paper's text, tables (after conversion to text), and images (after conversion to text) in two stages. In the first stage, we identify sections containing relevant data. In the second stage, we extract and integrate the data from these identified sections.

In the first stage, we use the following prompt to filter parts of all papers that contain the data of interest to the user:

\begin{tcolorbox}[breakable,title=Prompt for the First Stage of Data Extracting]

\textbf{\textcolor{blue}{System Prompt}}\\ 
You are an expert in the field of <INPUT1> and possess exceptional skills in analyzing whether the <INPUT\_TYPE>s in a paper contain data of interest. Your task is to evaluate the relevance of multiple <INPUT\_TYPE>s in the paper to the specified topic.

\textbf{\textcolor{blue}{Instructions}}\\ 
1. \textbf{Scope of Analysis}: The user will provide multiple first-level <INPUT\_TYPE>s, each of which may contain sub-<INPUT\_TYPE>s (second-level). Your focus should be solely on assessing the relevance of the first-level <INPUT\_TYPE>s to the topic. Ensure that the number of your responses matches the number of first-level <INPUT\_TYPE>s provided.

2. \textbf{Relevance Scoring}: For each first-level <INPUT\_TYPE>, determine its relevance to the topic of interest and assign a score between 0 and 1, where:

   - "0" indicates that the <INPUT\_TYPE> is \textbf{completely irrelevant}.
   
   - "1" indicates that the <INPUT\_TYPE> is \textbf{highly relevant}.
   
   - Scores \textbf{greater than 0.5} should be assigned if \textbf{any part of the <INPUT\_TYPE>} contains data relevant to the topic, even if only a small portion is relevant. This approach ensures that no potentially useful data is overlooked.
   
3. \textbf{Quality Emphasis}: Your primary objective is to maximize the identification of relevant <INPUT\_TYPE>s. Therefore, err on the side of inclusivity by assigning scores greater than 0.5 to any <INPUT\_TYPE> that contains even a minimal amount of relevant data. This strategy ensures comprehensive coverage and minimizes the risk of missing valuable information.

4. \textbf{Output Format}: Present your final assessment as \textbf{a list of scores} corresponding to each first-level <INPUT\_TYPE>. Do not include any additional text or explanations in your response.

\textbf{\textcolor{blue}{Example}}\\ 
$[0.8, 0.3, 0.9, 0.6, 0.7, ...]$

\end{tcolorbox}
\begin{figure}[ht]
    \vspace{0.01cm}
    \caption{\textbf{Prompt for the First Stage of Data Extracting.}}
    \label{fig:Prompt for the First Stage of Data Extracting}
\end{figure}

After the extraction in the first stage, we can remove redundant and irrelevant sections. Next, from all relevant data frames, we extract key data and consolidate it into a table. In the second stage, we use the following prompt:

\begin{tcolorbox}[breakable,title=Prompt for the Second Stage of Data Extracting]

\textbf{\textcolor{blue}{System Prompt}}\\ 
You are an expert in the field of <INPUT1> with a strong ability to organize and convert data from <INPUT\_TYPE>s. Your task is to transform all data from multiple <INPUT\_TYPE>s provided in a research paper into a new, integrated table based on a user-provided template. Your goal is to ensure that no data is left behind-every number, value, and piece of information from the original <INPUT\_TYPE>s must be included in the integrated table.

\textbf{\textcolor{blue}{Instructions}}\\ 
1. \textbf{Comprehensiveness}: Your primary objective is to include every single piece of data from the original <INPUT\_TYPE>s in the integrated table. Follow these rules to ensure complete data coverage:

   - \textbf{Transform all data from every <INPUT\_TYPE>}: Each <INPUT\_TYPE> contains valuable data, and you must transform every value, number, and data point from every <INPUT\_TYPE> into the integrated table. Do not skip any <INPUT\_TYPE>, row, column, or cell.
   
   - \textbf{Include all instances of data}: If data appears in multiple <INPUT\_TYPE>s (e.g., the same variable in <INPUT\_TYPE> 1 and <INPUT\_TYPE> 3), include all instances in the integrated table, even if they are identical, redundant, or similar.
   
   - \textbf{Include all statistical values}: If the original data contains statistical values (e.g., mean, maximum, minimum, range, percentiles, etc.), include all of these numbers in the integrated table. Do not omit any values, even if they appear repetitive or less significant.
   
   - \textbf{Include all case-specific values}: If the original data provides values for multiple cases (e.g., rainy season, dry season, different times, etc.), include all case-specific values in the integrated table. Do not omit any values, even if the case key is repeated across <INPUT\_TYPE>s.
   
   - \textbf{Include entire rows and columns}: If a row or column in a <INPUT\_TYPE> is relevant to the template, include all data points from that row or column in the integrated table. Do not leave out any values.
   
   - \textbf{Include entire <INPUT\_TYPE>s if relevant}: If all data in a <INPUT\_TYPE> is relevant, transform every data point into the integrated table as multiple rows. Do not exclude any part of the <INPUT\_TYPE>.
   
   - \textbf{Include incomplete data}: If any data is missing in the source <INPUT\_TYPE>s, use "NaN" as a placeholder in the integrated table. Every row does not need to be complete, but every piece of available data must be included.

2. \textbf{Single Integrated Table}: Provide only **one integrated table** in Markdown format. Do not create multiple tables.

3. \textbf{Table Format}: Format the integrated table according to the user-provided **Integrated Table Template**. Do not include the template itself in the output.

4. \textbf{Data Format}: Represent each numerical value as an integer or a float number. Exclude any symbols such as ">", "<", "~", "=", "+", "-", "(", ")", etc.

5. \textbf{Data Source Explanation}: After the integrated table, provide a clear explanation of the source of each data point. Specify the <INPUT\_TYPE> and the exact location (i-th row and j-th column) from which the data was transformed.

\textbf{\textcolor{blue}{Example}}\\ 
```markdown

| Column 1 | Column 2 |

|----------|----------|

| 10.5     | 20.3     |

| 15.2     | NaN      |

| 12.8     | 18.4     |

```

[The Start of Explanation]

1. The number 10.5: Comes from <INPUT\_TYPE> 2, Row 3, Column 2.

2. The number 20.3: Comes from <INPUT\_TYPE> 4, Row 5, Column 1.

3. ...

[The End of Explanation]

\end{tcolorbox}
\begin{figure}[ht]
    \vspace{0.01cm}
    \caption{\textbf{Prompt for the Second Stage of Data Extracting.}}
    \label{fig:Prompt for the Second Stage of Data Extracting}
\end{figure}

\section{Checker}

The checker is responsible for reviewing the data extraction and integration results from the data extractor. It scores the extraction results and provides feedback for modifications. If the score falls below a threshold, the integration results will be rejected, and revisions will be made based on the checker's suggestions. The prompt used in the checker is as follows:

\begin{tcolorbox}[breakable,title=Prompt for Checker]

\textbf{\textcolor{blue}{System Prompt}}\\ 
You are an expert in the field of <INPUT1>. Your task is to evaluate whether a student's table integration work is reasonable and accurate. 

Note that the student's objective is to comprehensively transform the data rather than simply extract it, so duplicated data or NaN values in the integrated table are normal and should not be penalized. 

Your focus should be on ensuring the student has included as much relevant data as possible from the source tables, rather than checking for duplicates or missing values caused by extraction.

\textbf{\textcolor{blue}{Instructions}}\\ 
Evaluate the submission based on the following three dimensions:

1. \textbf{Data Accuracy}: Whether the data in the integrated table exactly matches the original tables.

2. \textbf{Semantic Consistency}: Whether the data meanings remain consistent with the source tables.

3. \textbf{Data Completeness}: Whether maximum relevant data from source tables has been integrated.

Assessment rules:

1. \textbf{Dimensional Scoring}:

   - Evaluate each dimension independently (1-10 scale), highlighting strengths and weaknesses
   
   - Special note for "Data Completeness":
   
     - High score for integrating most/all data
     
     - For significant omissions, specify exactly where to find and include missing data

2. \textbf{Overall Score}:

   - Provide a composite 1-10 rating based on dimensional scores
   
   - Award minimum score for empty submissions

3. \textbf{Improvement Suggestions}:

   - Give concrete, actionable suggestions (using "You should..." phrasing)
   
   - Must precisely identify:
   
     - Location of missing data (e.g.: Column 3 in Table 2)
     
     - Data ranges needing inclusion (e.g.: Rows 5-10 in Table 1)
     
     - Specific fields requiring verification (e.g.: "Income" column in Table 3)

\textbf{\textcolor{blue}{Example}}\\ 
\{

'Data Accuracy': 9, 

'Semantic Consistency': 6, 

'Data Completeness': 8, 

'Overall Score': 7, 

'Suggestion': "You should add the data from Table 2, Column 3 to the integrated table, as it contains relevant information that is currently missing. Additionally, ensure that the values from Table 1, Rows 5 to 10 are included, as they have not been transferred. Finally, check Table 3, Column Revenue to confirm all its values are present in the integrated table."

\}

\end{tcolorbox}
\begin{figure}[ht]
    \vspace{0.01cm}
    \caption{\textbf{Prompt for Checker.}}
    \label{fig:Prompt for Checker}
\end{figure}

\section{Data Analyst}

The data analyst is responsible for analyzing the extracted data, such as classification, regression, and clustering. To achieve this, we automatically generate code for data analysis using LLMs in combination with data characteristics. The prompt used is as follows:

\begin{tcolorbox}[breakable,title=Prompt for Data Analyst]

\textbf{\textcolor{blue}{System Prompt}}\\ 
You are an expert in <INPUT1>. You are good at data analysis and visualization. Please complete the corresponding code according to user needs.

\textbf{\textcolor{blue}{Instructions}}\\ 
Now we have data, which is pandas data, where data.head() is as follows:

<INPUT1>

We have imported the following functions, you can freely choose and use them, and finally use plt.show() to display.\\

import numpy as np

import pandas as pd

from sklearn.linear\_model import LogisticRegression

from sklearn.svm import SVC

...

from sklearn.impute import SimpleImputer

import matplotlib.pyplot as plt\\

When replying, please strictly follow the following rules:

1. Please implement three visualization functions based on clustering, classification, and regression models respectively.

2. Do not use plt.savefig() function, use plt.show() function.

3. Use the plt.title() function to explain the meaning of the visualization. 
Please use a sentence to describe the meaning of the different coordinates or curves in the image so that the meaning of the image can be understood at first glance.
Please include appropriate line breaks in the title to avoid title overflow or obstruction.

4. All functions end with return.

\textbf{\textcolor{blue}{Example}}\\ 
def clustering(data):

    ...
    
    return\\

def classification(data):

    ...
    
    return\\

def regression(data):

    ...
    
    return\\

\end{tcolorbox}
\begin{figure}[ht]
    \vspace{0.01cm}
    \caption{\textbf{Prompt for Data Analyst.}}
    \label{fig:Prompt for Data Analyst}
\end{figure}

\section{Reporter}

After completing data extraction, integration, and analysis, the final stage is to summarize the findings into a meta-analysis report, combining data characteristics with the analysis results. The report is formatted in Markdown. We use the following prompt:

\begin{tcolorbox}[breakable,title=Prompt for Reporter]

\textbf{\textcolor{blue}{System Prompt}}\\ 
You are a senior Meta-Analysis expert in the field of <INPUT>. Please prepare a rigorous Meta-Analysis report based on the user's specified research direction, incorporating both structured tabular data and visual graphical data. The report must meet the following requirements:

\textbf{\textcolor{blue}{Instructions}}\\ 
1. \textbf{Data Integration and Analysis}

   - Conduct multidimensional statistical analysis of tabular data (mean/SD/effect size etc.)
   
   - Calculate heterogeneity (I**2) using standard methods (RevMan/STATA)
   
   - Provide professional interpretation of figures including:
   
     - Forest Plots
     
     - Funnel Plots
     
     - Sensitivity Analysis Plots

2. \textbf{Report Structure Standards}

   - Use Markdown formatting
   
   - Must embed figures using standard Markdown syntax:
   
     ```markdown
     
     ![Figure description](image\_URL\_or\_path)
     
     ```
     
   - Figures must be properly numbered (Figure 1, Figure 2 etc.)

3. \textbf{Content Requirements}

   - Results section must include:
   
     - Statistical test results from tables (p-values, CI intervals)
     
     - Key findings from figures
     
     - Subgroup analysis results (if applicable)
     
   - Discussion section must include:
   
     - Clinical/research implications
     
     - Publication bias assessment
     
     - GRADE evidence level

\textbf{\textcolor{blue}{Example}}\\ 
"""markdown

\# Meta-Analysis Report: XXX Research Direction\\

\#\# Methods

- Effect model selection (fixed/random effects)

- Heterogeneity testing methods\\

\#\# Results

\#\#\# Primary Outcomes

- Pooled effect size: OR 1.25 (95\%CI 1.10-1.42)

![Figure1: Forest plot of primary outcomes](path1)

\#\#\# Sensitivity Analysis

![Figure2: Leave-one-out sensitivity analysis](path2)\\

\#\# Discussion

- Comparison with existing literature

- Study limitations (e.g., included study quality)\\

\#\# References

 - List of references used in the report
 
"""

\end{tcolorbox}
\begin{figure}[ht]
    \vspace{0.01cm}
    \caption{\textbf{Prompt for Reporter.}}
    \label{fig:Prompt for Reporter}
\end{figure}

\end{document}